\pdfoutput=1

\documentclass[11pt]{article}

\usepackage[]{EMNLP2022}

\usepackage{times}
\usepackage{latexsym}

\usepackage[T1]{fontenc}

\usepackage[utf8]{inputenc}

\usepackage{microtype}

\usepackage{inconsolata}

\usepackage{graphicx}
\usepackage{float}
\usepackage{subfigure}

\usepackage{bm}
\usepackage{amsfonts}
\usepackage{booktabs}
\usepackage{multirow}
\usepackage{paralist}
\usepackage{CJKutf8}

\title{Extending Phrase Grounding with Pronouns in Visual Dialogues}

\author{
Panzhong Lu$^1$, Xin Zhang$^1$,
Meishan Zhang$^2$\thanks{~~Corresponding author.}~, Min Zhang$^2$ \\
$^1$School of New Media and Communication, Tianjin University \\
$^2$Institute of Computing and Intelligence, Harbin Institute of Technology (Shenzhen) \\
\texttt{\{panzhong171,hsinz\}@tju.edu.cn, \{zhangmeishan,zhangmin2021\}@hit.edu.cn} \\
}

\begin{document}
\begin{CJK}{UTF8}{gbsn}

\maketitle

\begin{abstract}
Conventional phrase grounding aims to localize noun phrases mentioned in a given caption to their corresponding image regions,
which has achieved great success recently.
Apparently, sole noun phrase grounding is not enough for cross-modal visual language understanding.
Here we extend the task by considering pronouns as well.
First, we construct a dataset of phrase grounding with both noun phrases and pronouns to image regions.
Based on the dataset, we test the performance of phrase grounding by using a state-of-the-art literature model of this line.
Then, we enhance the baseline grounding model with coreference information which should help our task potentially,
modeling the coreference structures with graph convolutional networks.
Experiments on our dataset, interestingly, show that pronouns are easier to ground than noun phrases,
where the possible reason might be that these pronouns are much less ambiguous.
Additionally, our final model with coreference information can significantly boost the grounding performance of both noun phrases and pronouns.
\end{abstract}

\section{Introduction}

Grounded language learning has been prevailing for decades in many fields  \citep{chandu-etal-2021-grounding}, generally aiming to learn the real-world meaning of textual units (e.g., words or phrases) by conjointly leveraging the perception data (e.g., images or videos).
\citet{bisk-etal-2020-experience} advocate that we cannot overlook the physical world that language describes when doing language understanding research from a novel perspective.
In particular, with the stimulation of modeling techniques and multi-modal data collection paradigms, the task has made excellent progress in the downstream tasks, which involves multi-modal question answering \citep{DBLP:journals/ijcv/AgrawalLAMZPB17, chang2022webqa}, video-text alignment \citep{DBLP:conf/iccv/YangBG21} and robot navigation \citep{roman-roman-etal-2020-rmm, gu-etal-2022-vision}.

\begin{figure}
    \centering
    \includegraphics[width=0.47\textwidth]{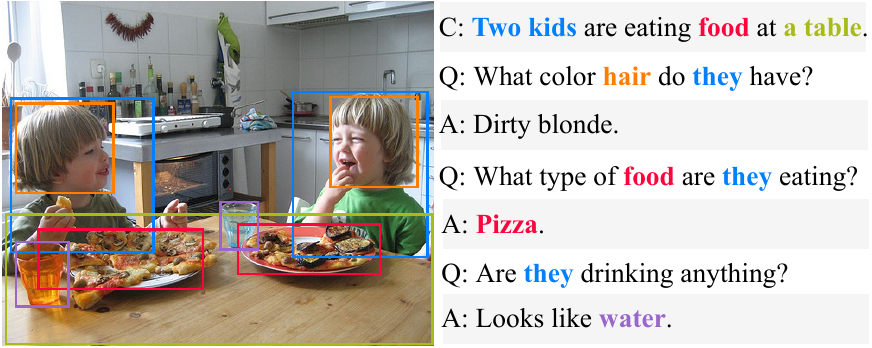}
    \caption{An example of grounding noun phrases and pronouns referred in the caption and dialogue (partly) to the associated image regions. With an image described by a caption, two people are discussing what they can see. Here, we annotate the same object with the same color, respectively. And obviously, the same object mentioned in the text naturally forms a coreference chain.}
    \label{fig:dataset}
\end{figure}

Typically, as one branch of grounded language learning, phrase grounding, first proposed by \citet{DBLP:conf/iccv/PlummerWCCHL15}, also plays a key role in visual language understanding. 
Its goal is to ground the phrases in a given caption to the corresponding image regions.
Recently, many researchers have attempted varied approaches to explore this task.
\citet{DBLP:conf/aaai/MuTTYZ21} propose a novel graph learning framework for phrase grounding to distinguish the diversity of context among phrases and image regions.
\citet{wang-etal-2020-maf} develop a multimodal alignment framework to utilize the caption-image datasets under weak supervision.
\citet{DBLP:conf/iccv/KamathSLSMC21} advance phrase grounding with their end-to-end modulated pre-trained network named MDETR.
Overall, the natural language processing (NLP) and computer vision (CV) communities have seen huge achievements in the task of phrase grounding.

In spite of its apparent success, there remains a worth-thinking weakness.
Almost all previous works mainly focus on the noun phrases/words, which can derive their meanings by the expressional forms to some extent.
There is little work that takes account into pronouns.
As shown in Figure \ref{fig:dataset},
pronouns definitely have underlying effects on the performance of visual grounding,
which should be carefully examined \citep{yu-etal-2019-see}.
As a result, here we shift our eyes from the common (almost noun) phrase grounding with the extension of pronouns for the first time.

In this paper, we present the first work for investigating phrase grounding that includes pronouns, and explore how coreference chains can have an effect on the performance of our task.
We annotate an initial dataset based on visual dialogue \citep{VisDialog}, as shown in Figure \ref{fig:dataset}.
For the model, we can directly apply MDETR \citep{DBLP:conf/iccv/KamathSLSMC21}, which is an end-to-end modulated detector.
However, the model does offer much information to understand pronouns.
Thus, we enhance the vanilla model with coreference information from the dialogue end, where a graph neural network is adopted to encode the graph-style coreference knowledge.

Finally, we conduct experiments on our constructed dataset to benchmark the extended phrase grounding task.
According to the results, we find that interestingly, pronouns are easier to ground by MDETR than phrases.
The underlying reason might be that the pronouns are always more important during dialogue, leading to less ambiguity in speech communication.
In addition, our final model can be significantly enhanced by adding the gold graph-style coreference knowledge;
however, the model fails to obtain any positive gain when the coreference information is sourced from a state-of-the-art machine learning model.
We conduct several in-depth analyses for comprehensive understanding of our task as well as the model.

In summary, our contributions are as follows:
\begin{compactitem}
  \item We extend the task of phrase grounding by taking account of pronouns, and correspondingly establish a new dataset manually, named VD-Ref, which is the first dataset with ground-truth mappings from both noun phrases and pronouns to image regions.
  \item We benchmark the extended phrase grounding task by a state-of-the-art model, and also investigate our task with the coreference knowledge of the text, which should benefit our task straightforwardly.
  \item We observe several unexpected results by our empirical verification, and to understand these results, in-depth analyses are offered to illustrate them, which might be useful for the future investigation of phrase grounding.
\end{compactitem}

\section{Our Task and The \texttt{VD-Ref} Dataset}

\begin{table}[tb]
  \centering
  \resizebox{0.48\textwidth}{!}{
  \begin{tabular}{crrrrr}
  \toprule
  Sect.  & \#Img & \#Pronoun & \#Phrase & \#Box & \#Coref \\
  \midrule
  Train & 6199 & 18600 & 35118 & 16559 & 14582 \\
  Dev   & 1063 & 3256 & 5739 & 3074 & 2503 \\
  Test  & 1595 & 4033 & 7941 & 4347 & 3754 \\
  \bottomrule
  \end{tabular}
  }
  \caption{
  Data statistics of our constructed dataset. \#Box means the number of bounding boxes in the image. \#Coref means the number of coreference chains.
  }
  \label{table:dataset}
\end{table}

\subsection{Task Description}
The phrase grounding task's general purpose is to map multiple noun phrases to the image regions, however, in this paper, we take the challenge a step further by grounding various noun phrases and pronouns from the given dialogue to the appropriate regions of an image. Take Figure \ref{fig:dataset} for example, with all the expressions mentioned in the dialogue, like the coreference chain that includes ``Two kids'' and ``they'', the task needs to predict the corresponding regions of the object ``kids'' using bounding boxes in image.

Formally, we define the task as follows: given an image $\I$ and the corresponding ground-truth dialogue $D$, we denote $M = \left\{ N, P \right\}$ as all the language expressions, typically, $N$ is the oun phrases and $P$ is the pronouns, the prime objective of the task is to predict a bounding box (or bounding boxes) $B$ for each expression.

\subsection{Data Collection}
With the aim to build a high-quality dataset that includes sufficient pronouns, we adopt the large-scaled VisDialog dataset \citep{VisDialog} which contains 120k images from the COCO \citep{DBLP:conf/eccv/LinMBHPRDZ14}, where each image is associated with a dialogue\footnote{If not specified, the following dialogues that are discussed all contain a caption.} around to the image.
We randomly choose a set of 10k complete sets from the VisDialog dataset,
and use the StanfordCoreNLP \citep{manning-etal-2014-stanford} tool to tokenize the sentences, making it proper for the succeeding human annotation.

\subsection{Annotation Process}

The whole annotation workflow is divided into three stages as follows: (\romannumeral1) developing a convenient online tool for the user annotation; (\romannumeral2)  setting up a standard annotation guideline according to our task purpose;
(\romannumeral3)  Recruiting sufficient expert users to annotate the dataset and ensuring each instance with three annotations.
Firstly, we adopt the label-studio platform \cite{Label/Studio} as the basis to design a user-friend interface targeted to our task, where the concreted interface is shown in Appendix \ref{sec:appendix:interface}.
Then, we let three people with the visual grounding research experience previously as our experts.
They annotate 100 data-pairs together as examples, and establish an annotation guideline based on their consensus after several discussions.

Next, we recruit a number of college students who are expertised at English skills to annotate our dataset. Before starting our task, the students are asked to read the guideline of the annotation process carefully and attempt to annotate some test sets of data, during this period, we examine these students and choose 20 of them to do the following annotation task. 
In the annotation of each datapoint, the prepared data is split into micro-tasks so that each one consists of 500 dialogues.
We assign three workers to each micro-task, and their identities are remained hidden from each other.
After all annotation tasks are finished, we let our experts check the results and make corrections of the unconsistent annotations as well.

Finally, we establish the \texttt{VD-Ref} dataset, which is annotated manually with the noun phrases and pronouns that naturally form the coreference chains as well as the relevant bounding boxes in images.

\begin{figure}[tb]
  \centering
  \includegraphics[width=0.48\textwidth]{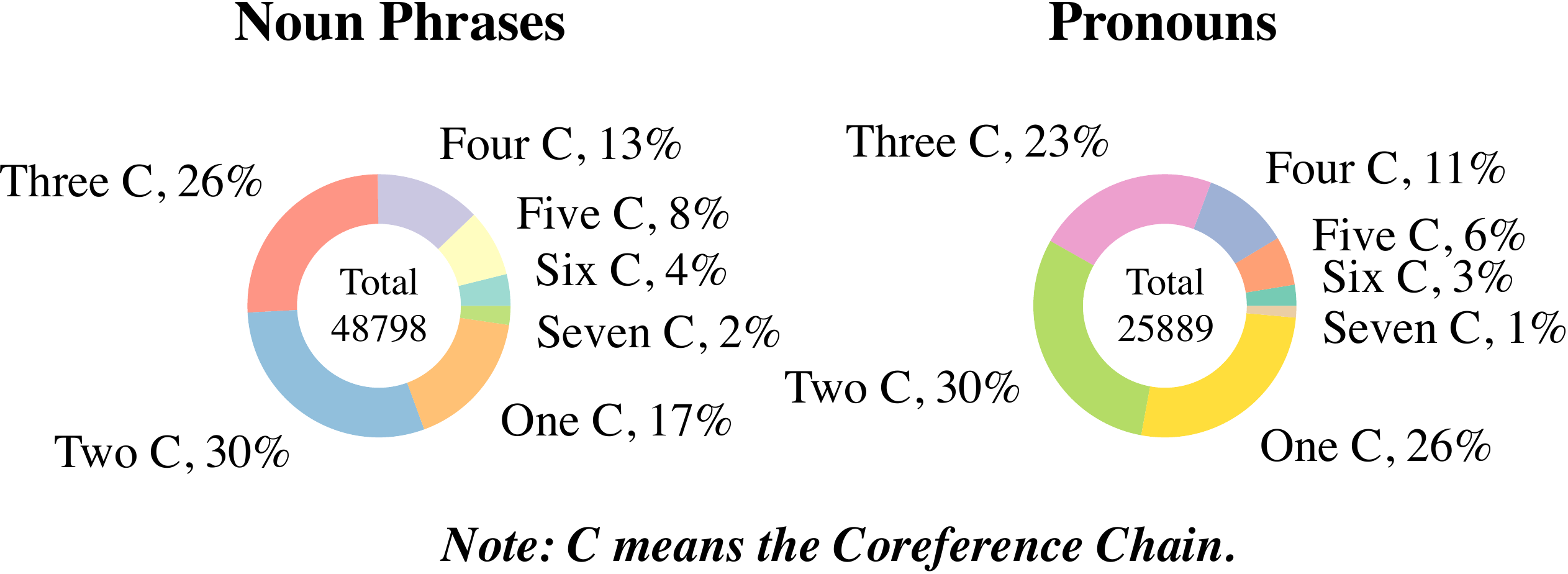}
  \caption{The proportion of noun phrases and pronouns in different number coreference chains.}
  \label{fig:pie}
\end{figure}

\subsection{Statistics of the \texttt{VD-Ref} Dataset}

Totally, we collect 74,687 entity mentions and 23,980 objects from 8,857 VisDialog datasets, where the mentions include 48,798 noun phrases and 25,889 pronouns, on average, a dialogue consists of 5.51 noun phrases and 2.92 pronouns. On the contrary, the existing datasets for phrase grounding hardly consider the pronouns. The ReferItGame dataset \citep{kazemzadeh-etal-2014-referitgame} only involves in the noun and noun phrases, while the Flickr30k Entities dataset does not label the corresponding bounding boxes in images, although it annotate the pronouns in captions.

Alternatively, because of the diversity of our dataset, the number of coreference chains varies. As Figure \ref{fig:pie} shows, the pie charts display the distinctive distributions of noun phrases and pronouns in the \texttt{VD-Ref} dataset. It is clear that whether for the noun phrases or the pronouns,  
the dialogues that have no more than three coreference chains account for the major proportion, 
up to 70\%, accordingly,
the dialogues that have more than three coreference chains constitute the rest proportion.

\begin{figure*}\centering
\includegraphics[scale=0.65]{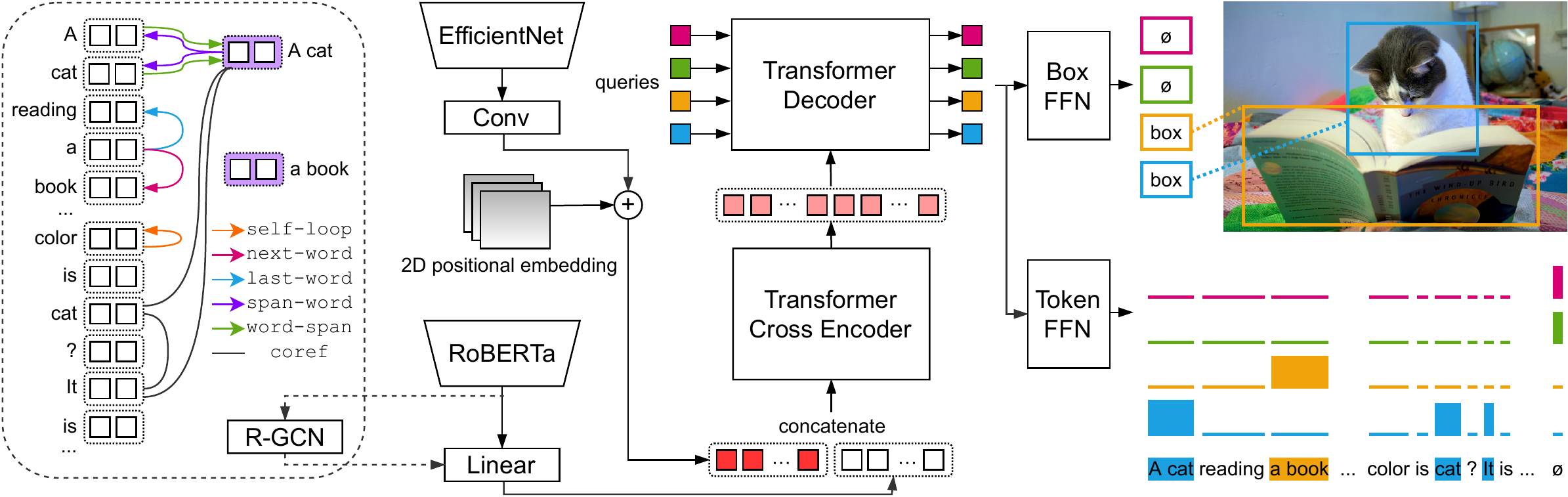}
\caption{MDETR Model (right) and our suggested coreference graph encoding (left, dashed). Here, we use R-GCN encode the coreference graph into the roberta representation and directly fed the output into the linear layer.}
\label{fig:model}
\end{figure*}

Moreover, as the mentions of the coreference chains and bounding boxes come in pairs, we can define the coreference chain into four types:

\begin{compactitem}
  \item \textbf{one mention vs. one box}: This type contains only one mention and one corresponding box, indicating that the chain exclude pronoun.
  \item \textbf{one mention vs. boxes}: As the referred object is separated into several regions, more than one box is needed to annotate.
  \item \textbf{mentions vs. one box}: In this coreference chain, all noun phrases and pronouns refer to the same single box on the image.
  \item \textbf{mentions vs. boxes}: This type contains several mentions that have noun phrases and pronouns and associated multiple boxes.
\end{compactitem}

Finally, the train, validation and test sets contain 6,199 (70.00\%), 1,063 (12.00\%) and 1,595 (18.00\%) image-dialogue pairs, respectively. We report other statistics in Table \ref{table:dataset} as well.

\section{Method}
Recent works \citep{DBLP:conf/iccv/KamathSLSMC21,DBLP:conf/cvpr/li-etal-glip} bring the successful vision-language transformer architecture and the pre-train-then-finetune paradigm to the phrase grounding task,
achieving state-of-the-art performance.
To explore our constructed dataset, we adopt the representative MDETR \citep{DBLP:conf/iccv/KamathSLSMC21} model.
Meanwhile, we propose to enhance the textual representations with the natural coreference chains in texts by Relational Graph Convolutional Networks (R-GCN) \citep{DBLP:conf/esws/SchlichtkrullKB18}.
Bellow, we briefly describe how MDETR learns and grounds, and then present our suggested coreference graph encoding.

\subsection{Grounding Model}
As depicted in Figure \ref{fig:model}, for a given image-text pair, MDETR first use an image-encoder \citep[EfficientNet]{DBLP:conf/icml/TanL19} to extract visual features.
Then, the features are projected to the image-text shared embedding space by a conv layer, flattened to a sequence, and added with 2-D positional embeddings.
Similiarly, a text-encoder \citep[Roberta]{DBLP:journals/corr/abs-1907-11692} and a linear layer are used to extract and project textual features, respectively.
Next, we concatenate vectors of two modalities into one sequence, encoding it by a transformer \citep{DBLP:conf/nips/VaswaniSPUJGKP17} encoder.
We set $N$ queries and apply a transformer decoder to cross attend the encoded sequence.

Finally, for each one of $N$ hidden states from the decoder, two feedforward networks (FFN) separately predict the object box and a distribution over all 
\emph{token positions} that correspond to the object, which named {\bf soft token prediction}.
Figure \ref{fig:model} shows an example that a query predict the box of the cat and a distribution where tokens refer to this cat are with highest values.

\paragraph{Training.}
MDETR uses the bipartite matching \citep{DBLP:conf/eccv/CarionMSUKZ20,DBLP:conf/ijcai/Tan0Z0Z21} to find the best match between the predicted boxes and the gold-standard objects then computes the box losses (L1 \& GIoU).
The soft token prediction is supervised by a soft-cross-entropy between the predicted distribution and a uniform distribution, where tokens refered to the matched gold box have equal probabilities and sum to $1$.
In addition, the matching cost consists of this grounding loss and the box L1 \& GIoU losses.
The final loss for the MDETR training is the weighted sum of the above losses and a extra contrastive alignment loss\footnote{
This loss is able to align the query hidden state from the decoder and it's matched referring tokens, please refer to the §2.2.2 of \citeposs{DBLP:conf/iccv/KamathSLSMC21} paper for more details.
}.

\paragraph{Inference.}
For each referring expression, we rank all $N$ proposed boxes by scores of the max over scores assigned to the tokens in this expression, and output the top $10$ boxes for the evaluation.

\subsection{Coreference Graph Encoding}
By carefully examining the MDETR model in our extended task, we find that it actually predicts the coreferenced expressions for each detected object to some extend.
We guess that explicitly injecting the text coreference information into the representations could boost the model performance to some extent.
Thus we propose to encode a simple coreference graph via R-GCN.

\paragraph{Graph Construction.}
Following the previous graph-based NLP studies \citep{sahu-etal-2019-inter,DBLP:journals/corr/abs-2106-06090,hu-etal-2020-selfore,hu-etal-2021-gradient}, we construct our coreference graph in two steps.
For the node building, we first initiate the \texttt{word} nodes by the input text embeddings.
To represent the multi-word mention in text, we generate a virtual \texttt{span} node\footnote{
We offer an ablation study (§\ref{sec:ablation}) to verify the effectiveness of this scheme.
} and setup the embedding by the mean embedding of all words in it.

Based on the above two node types (i.e., \texttt{word} \& \texttt{span}), we build the coreference graph with the following six edge types:
\begin{compactitem}
    \item \texttt{self-loop}: include the information of itself.
    \item \texttt{next-word}: to keep the sequential information, we link a word to its next word.
    \item \texttt{last-word}: likewise link a word to the last.
    \item \texttt{span-word}: we link a span node to its words for the graph message passing.
    \item \texttt{word-span}: likewise link a word to its span.
    \item \texttt{coref}: we use this undirected edge to connect words or spans refered to the same object.
\end{compactitem}

\begin{table*}
\centering
\resizebox{0.98\textwidth}{!}{
\begin{tabular}{lc|ccc|ccc|ccc}
\toprule[1.25pt]
& Coref & \multicolumn{3}{c|}{Recall@1} & \multicolumn{3}{c|}{Recall@5} & \multicolumn{3}{c}{Recall@10} \\
Model & F1 & Overall & Pronoun & Phrase   & Overall & Pronoun & Phrase   & Overall & Pronoun & Phrase \\
\midrule
\multicolumn{11}{c}{\textsc{Any-Box-Protocol}} \\
\midrule
MDETR                & -     & 43.35 & 50.15 & 39.94   & 57.18 & 67.35 & 52.13   & 65.04 & 75.41 & 60.29 \\
MDETR + NeuralCoref  & 37.6  & 42.04 & 49.39 & 38.36   & 53.72 & 63.50 & 48.84   & 61.91 & 72.60 & 56.60 \\
MDETR + C2f-SpanBERT & 66.0  & 42.36 & 49.45 & 38.87   & 54.80 & 64.07 & 50.28   & 63.58 & 73.48 & 58.61 \\
MDETR + Gold$^\dagger$ & 100  & \bf 47.54 & \bf 58.79 & \bf 41.91   &\bf  59.30 &\bf  70.52 &\bf  53.69   &\bf  66.67 &\bf  76.83 &\bf  61.59 \\
\midrule
\multicolumn{11}{c}{\textsc{Merged-Box-Protocol}} \\
\midrule
MDETR                & -     & 51.98 & 60.86 & 47.59   & 62.86 & 71.98 & 58.40   & 68.03 & 77.20 & 63.61 \\
MDETR + NeuralCoref  & 37.6  & 51.04 & 62.14 & 45.60   & 62.01 & 72.45 & 56.79   & 67.03 & 77.03 & 61.99 \\
MDETR + C2f-SpanBERT & 66.0  & 51.96 & 62.45 & 46.71   & 62.44 & 72.46 & 57.43   & 67.55 & 76.90 & 62.33 \\
MDETR + Gold$^\dagger$ & 100  &\bf  55.43 &\bf  65.86 &\bf  50.26  &\bf  65.72 &\bf  74.73 &\bf  61.23   &\bf  70.52 &\bf  78.64 &\bf  66.50 \\
\bottomrule[1.25pt]
\end{tabular}
}
\caption{
Test results. $^\dagger$ means the result is statistically significant compared with MDETR.
}\label{table:result}
\end{table*}

\paragraph{R-GCN Encoding.} We compute node representations on this edge-labeled graph by the R-GCN \citep{DBLP:conf/esws/SchlichtkrullKB18}.
Formally, we denote the hidden representation of node $i$ in the $l$-th R-GCN layer as $\bm{x}_i^{(l)} \in \mathbb{R}^{d^{(l)}}$, where $d^{(l)}$ is the hidden dimension. The message-passing framework of R-GCN is defined as follows:

{\small
\begin{equation}
\bm{x}_i^{(l+1)} = \sigma \left( \sum_{r \in \mathcal{R}} \sum_{j \in \mathcal{N}_i^r} \frac{1}{|\mathcal{N}_i^r|} \bm{W}_r^{(l)}\bm{x}_j^{(l)} + \bm{W}_0^{(l)}\bm{x}_i^{(l)} \right),
\end{equation}
}where $\mathcal{R}$ denotes the relation set except the \texttt{self-loop}.
$\mathcal{N}_i^r$ is the set of neighbouring nodes under the relation $r \in \mathcal{R}$.
$\bm{W}_r$ is the feature transformation matrix for relation $r$, while $\bm{W}_o$ corresponds to the \texttt{self-loop}.

In the end, we re-construct the sequence from all \texttt{word} node representations of the last layer.
It is worth to note that this coreference graph encoding is general and could be applied to any grounding models.
In our experiments, the output of the R-GCN is directly fed to the Linear layer.

\section{Experiment}

\subsection{Settings}

\paragraph{Implementation.}
We use the pretrained MDETR with Roberta-base and EfficientNet-B3.\footnote{
\url{https://github.com/ashkamath/mdetr}
}
We employ 2-layer R-GCN\footnote{
We use the R-GCN implemetation from PyTorch Geometric \citep{Fey/Lenssen/2019}.
} to encode the Roberta representations.
We use the AdamW \citep{DBLP:conf/iclr/LoshchilovH19} to update model parameters with lr $1e^{-5}$, weight decay $1e^{-4}$, and batch size 16.
The lr of the re-initiated MDETR soft token prediction head and R-GCN module is set to $1e^{-4}$.
The 2-norms of gradients are clipped to a maximum of $0.1$ to avoid the gradient explosion problem.
All experiments are implemented with AllenNLP \citep{gardner-etal-2018-allennlp} and conducted on a RTX 3090 GPU.

\paragraph{Coreference.}
We consider three ways to obtain coreference chains for our graph-encoding:
\begin{compactitem}
\item \texttt{Gold}: the gold-standard coreference chains annotated in our dataset.
\item \texttt{NeuralCoref}:
the off-the-shelf coreference resolution toolbox based-on SpaCy from \citet{neuralcoref}, we load the ``en-core-web-md'' model for SpaCy.
\item \texttt{C2f-SpanBERT}:
the widely used span-based coarse-to-fine model \citep{lee-etal-2018-higher} with a pretrained SpanBERT-large-cased \citep{joshi-etal-2020-spanbert}.\footnote{
\url{https://github.com/allenai/allennlp-models}
}
We train it with the gold coreferences and perform an 5-fold cross-validation to get predictions of the whole dataset.
\end{compactitem}
In our main results, we train the MDETR + \texttt{NeuralCoref} or \texttt{C2f-SpanBERT} with only pseudo coreferences, which would fit the real scenario. We will investigate the recent state-of-the-art works of text coreference resolution\cite{wu-etal-2020-corefqa} and update the results in the future version paper.

\begin{table*}
\centering
\resizebox{0.98\textwidth}{!}{
\begin{tabular}{lcc|ccc|ccc|ccc}
\toprule[1.25pt]
& Coref & & \multicolumn{3}{c|}{Miss} & \multicolumn{3}{c|}{Part} & \multicolumn{3}{c}{Correct} \\
Model & F1 & Protocol & Overall & Pronoun & Phrase   & Overall & Pronoun & Phrase   & Overall & Pronoun & Phrase \\
\midrule
\multicolumn{3}{c|}{\#Mention} && 498 & 5507  && 3335 & 2107  && 125 & 300 \\
\multirow{2}{*}{+ NeuralCoref} & \multirow{2}{*}{37.6} & Any-Box   & 34.52 & 45.78 & 33.50   & 48.44 & 48.19 & 48.84   & 39.76 & 45.60 & 37.33 \\
&& Merged-Box  & 42.10 & 54.22 & 41.00   & 60.69 & 63.27 & 56.62   & 47.29 & 64.00 & 40.33 \\
\midrule
\multicolumn{3}{c|}{\#Mention} && 410 & 2921  && 2497 & 3452  && 1051 & 1541 \\
\multirow{2}{*}{+ C2f-SpanBERT} & \multirow{2}{*}{66.0} & Any-Box   & 26.06 & 45.12 & 23.38   & 47.96 & 47.94 & 47.97    & 45.99 & 46.72 & 45.49 \\
&& Merged-Box    & 32.81 & 49.02 & 30.54    & 60.30 & 63.84 & 57.73    & 57.48 & 64.41 & 52.76 \\
\bottomrule[1.25pt]
\end{tabular}
}
\caption{
Test recall@1 of MDETR with \texttt{NeuralCoref} or \texttt{C2f-SpanBERT}, by the mention prediction types, where \texttt{Miss} means a mention is not extracted by coref models, \texttt{Part} (resp. \texttt{Correct}) denote a mention is extracted with the incorrect (resp. correct) coreference cluster. We also provide the number of each type, i.e., the \#Mention row.
}\label{table:autokind}
\end{table*}

\paragraph{Evaluation.}
Following previous studies, we compute the Recall@$k$ to measure whether the model is able to give the ``correct'' box in top $k$ predictions, where a box is treated as ``correct'' if the Intersection-over-Union (IoU) between it and a ground-truth box is above a threshold of $0.5$.
For each text mention, we consider $k \in \{ 1, 5, 10 \}$.
We conduct experiments on both \texttt{Any-Box} and \texttt{Merged-Box} protocol, where the former decides a proposed box is correct to a mention when it has an Iou $> 0.5$ with \emph{any} of the gold boxes of this mention, while the latter merges all ground-truth boxes of a mention into one smallest enclosing box.

We use the best-performing model on the devset to evaluate the performance of the testset.
We run each setting by 5 different random seeds, and the average test scores are reported.
We regard a result as statistically significant when the p-value is below $0.05$ by the paired t-test with baseline MDETR.

\begin{figure}[tb]
  \centering
  \includegraphics[scale=0.5]{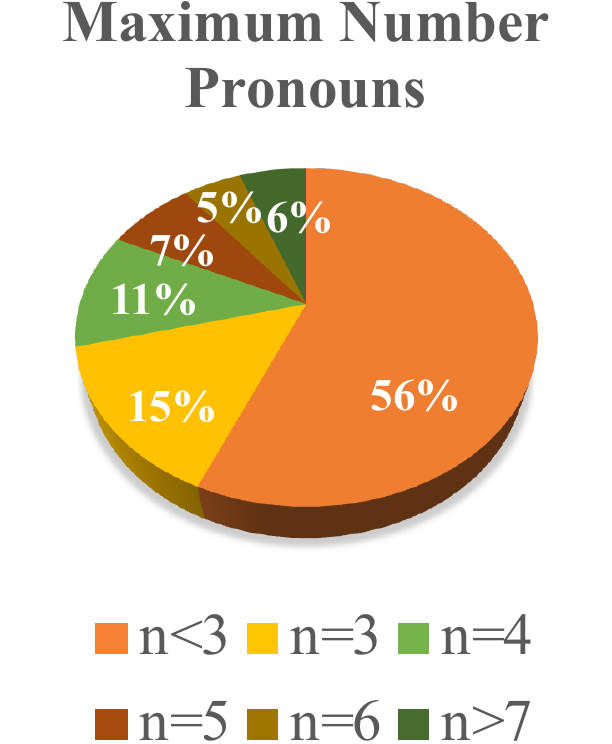}
  \hspace{0.1in}
  \includegraphics[scale=0.75]{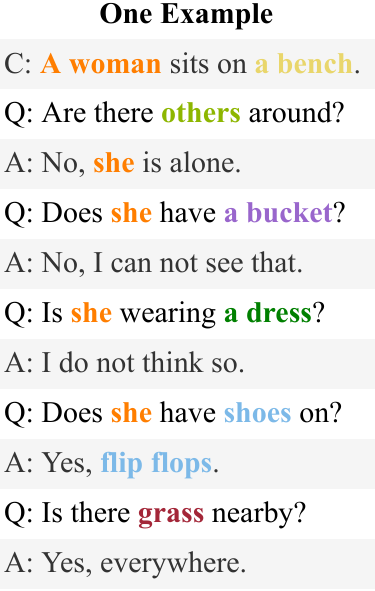}
  \caption{The assessment of our whole dataset on the maximum number of pronouns (in one coreference chain) for every dialogue(left) and one example (right).}
  \label{fig:pronoun}
\end{figure}

\subsection{Main Results}\label{sec:mainresult}

Table \ref{table:result} summarizes the main results of phrase grounding experiments on our VD-Ref dataset. We group the models into two settings, \texttt{Any-Box} and \texttt{Merged-Box} protocols, and report the performance of grounding pronouns and phrases in terms of Recall@$k$ $(k=1,5,10)$.
In details, we have the following intriguing findings:

Among all the models, we find that pronouns are easier to ground than phrases, no matter the protocol setting.
The possible reason is that as an essential part of the sentence in dialogue, pronouns are straightforward and appear more frequently, containing richer details in context, thus promoting the performance to be grounded.

Besides, after comparing the results of the MDETR with gold (MDETR + Gold) and without (MDETR), we see that adding the gold graph-style coreference knowledge can also considerably improve the model's performance.
This empirically supports the value of introducing coreference knowledge.
Noticeably, the Recall@$10$ is generally utilized to evaluate the best recall performance, and at this point, MDETR would reach its limit on this task, making it hard to be improved to some extent, while the addition of the gold graph-style coreference increase that by more than 1\%, which further proof the significance of coreference knowledge.

However, we still observe that performance declines when we apply machine learning models (e.g., \texttt{NeuralCoref} and \texttt{C2f-SpanBERT})  to obtain the coreference chains for our graph structure representations.
One possible reason is that these models do not do so  well in dialogues, making investigating the more thorough sense worthwhile.

\subsection{Analysis}

\paragraph{Pronouns outperform phrases.}
To dig into the in-deep reasons for this performance, 
we count the maximum number of pronouns (in one coreference chain) for every dialogue, and select one annotated dialogue as an example (see Figure \ref{fig:pronoun}). Here, we find that pronouns frequently occur in dialogues, and the maximum number of pronouns larger than three accounts for 44\% in our dataset, indicating the importance of pronouns in dialogues. Besides, take Figure \ref{fig:pronoun} (right) for an example. Four pronouns refer to ``woman'' in the dialogue. The reason behind this is that expressing pronouns are more concise to refer to the specific object, reducing ambiguity in communication.

\paragraph{Detailed Comparison of Non-Gold Methods.}
To find reasons for the unexpected failure of MDETR with pseudo coreferences from \texttt{NeuralCoref} and \texttt{C2f-SpanBERT}, 
we split testset mentions by the coreference cluster prediction of each of them is failed/partially correct/correct.
Detailed R@1 values in the three types are in Table~\ref{table:autokind}.
When the prediction fails (\texttt{Miss}), model performances are significantly lower than the average, which hurts the overall performances much since these mentions took considerable portions.
Surprisingly, the partially correct scores are above the average, which means that even with the defective coreference knowledge, models could precisely ground to a certain extent.
Improving the coref model recall could be an effective way to promote grounding performance of suggested method.

\begin{figure}[tb] 
  \centering
  \includegraphics[width=0.48\textwidth]{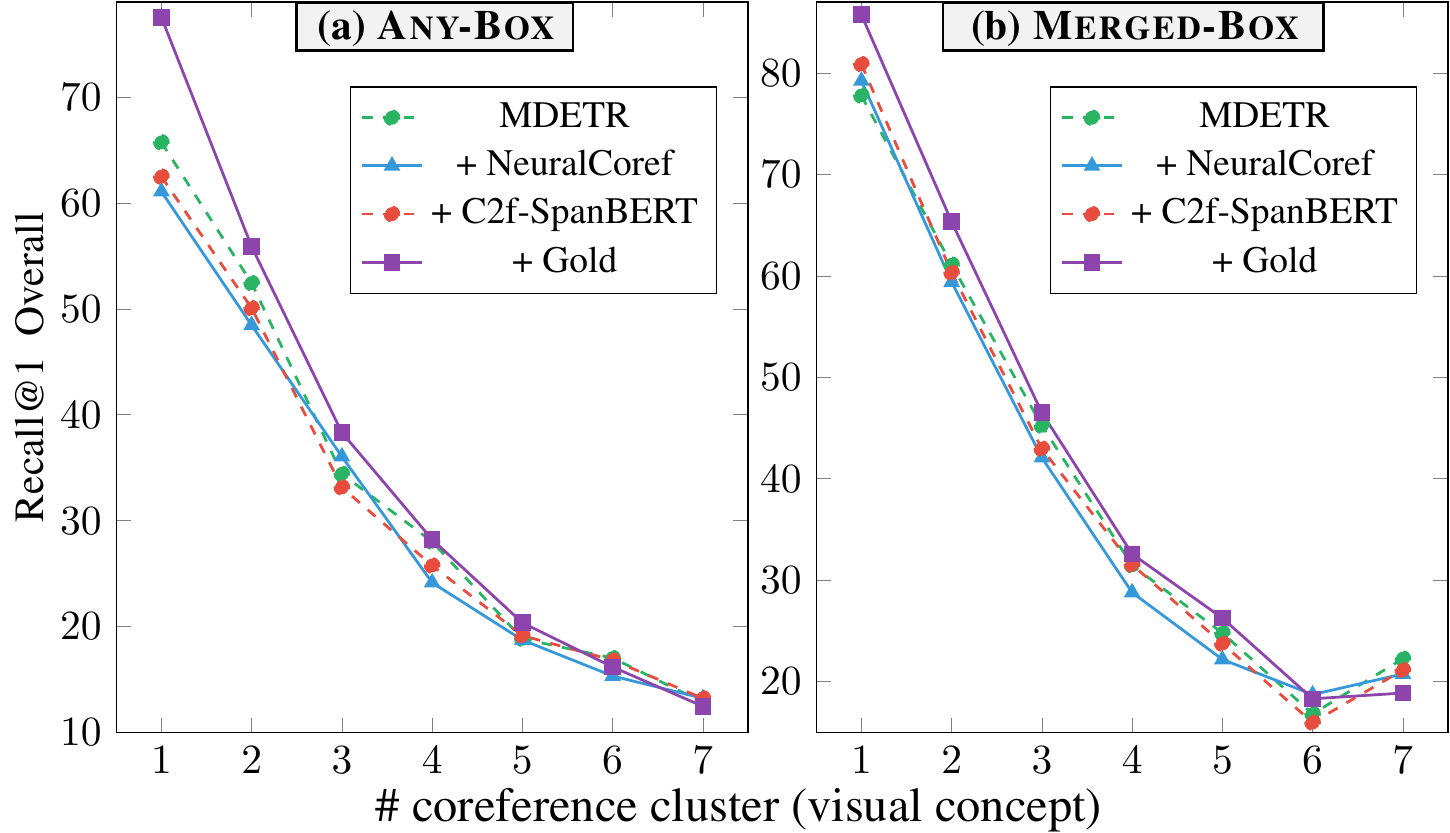}
  \caption{
The test Recall@1 (overall) scores grouped by \#cluster, which act as the number of visual concepts and represent the difficulty of a data point.
}
  \label{fig:coref-cluster}
\end{figure}

\paragraph{Understanding Complex Scenarios.}
Generally, models would perform worse in a complex scene than in a simple one.
We design analysis to validate it in practice to evaluate model abilities in complex scenarios.
We measure the difficulty of a scenario (data point) by the number of \emph{coreference clusters}, which represents the number of \emph{visual concepts} that need identifying, grouping testset into different parts.
As shown in Figure \ref{fig:coref-cluster}, performances of all methods decline as the clusters increase.
The \texttt{Gold} offers notable improvements in the simple data (\#cluster $\le 3$).
All methods perform poorly in complex scenarios, which would be one major limitation of phrase grounding models currently.

\begin{figure}[tb] 
  \centering
  \includegraphics[width=0.48\textwidth]{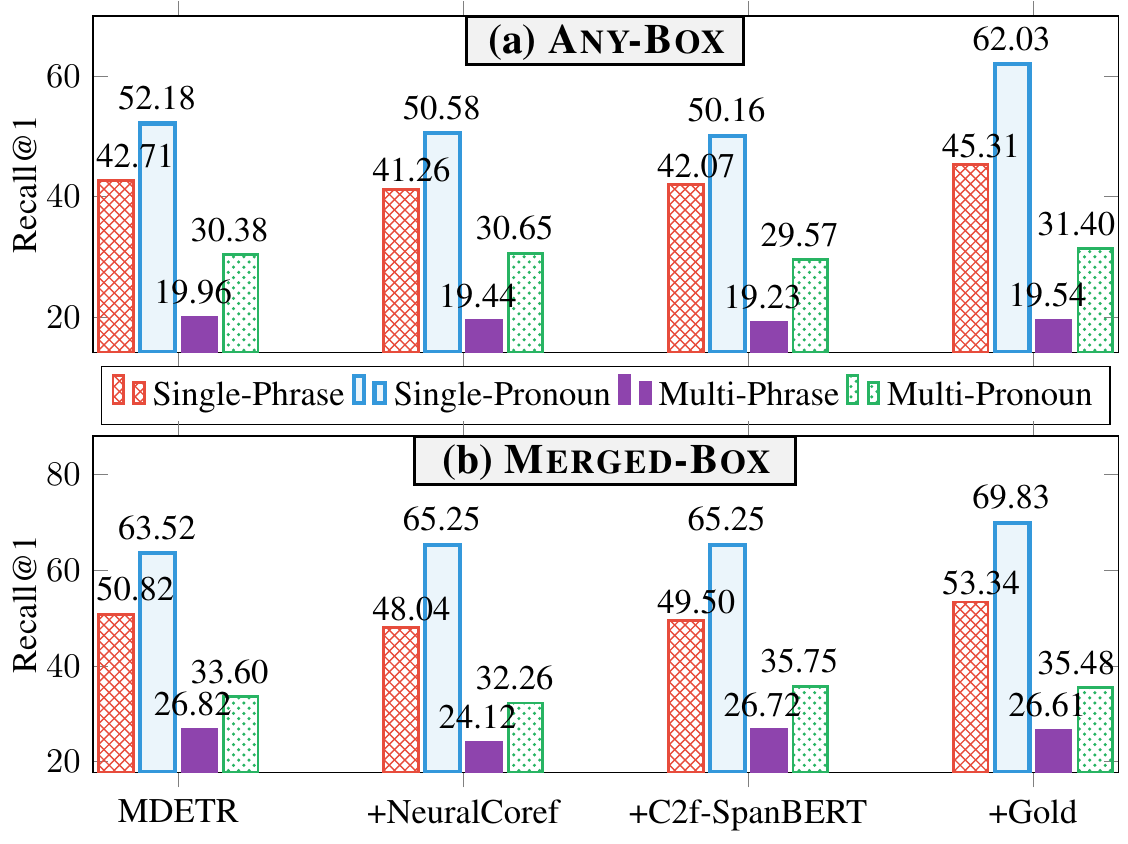}
  \caption{
The test Recall@1 of one phrase/pronoun referring to Single/Multi visual objects.
}
  \label{fig:single-multi}
\end{figure}

\paragraph{Grounding Single/Multi-Object Mentions.}
As discussed by \citet{DBLP:conf/iccv/KamathSLSMC21}, the \texttt{Any-Box} and \texttt{Merged-Box} protocols are used to handle that the recall@$k$ implies each mention referring to single object.
Here we divide mentions into two types, single and multi reference (e.g., the multi reference ``two kids'' referred two boxes in Figure~\ref{fig:dataset}),
and compare the performances.
In Figure~\ref{fig:single-multi}, indeed the multiple reference are much challenging, showing shocking gaps to the single.
That is, except for the challenges in complex scenarios (instance-level), the model ability on multi-object mentions (prediction-level) also need to be upgraded.
Besides, the performance of pronouns is consistently better than that of noun phrases as expected.

\begin{table}
\centering
\resizebox{0.44\textwidth}{!}{
\begin{tabular}{llccc}
\toprule[1.25pt]
Objective & Model & R@1 & R@5 & R@10 \\
\midrule
\multicolumn{5}{c}{\textsc{Any-Box-Protocol}} \\
\midrule
\multirow{2}{*}{Phrase} & MDETR  & 38.98 & 52.01 & 60.43 \\
& MDETR + Gold                  & 41.43 & 53.15 & 61.54 \\
\midrule
\multirow{2}{*}{Pronoun} & MDETR & 59.98 & 69.77 & 74.93 \\
& MDETR + Gold                  & 62.31 & 70.51 & 75.71 \\
\midrule
\multicolumn{5}{c}{\textsc{Merged-Box-Protocol}} \\
\midrule
\multirow{2}{*}{Phrase} & MDETR  & 47.34 & 57.75 & 63.60 \\
& MDETR + Gold                  & 48.48 & 58.03 & 63.89 \\
\midrule
\multirow{2}{*}{Pronoun} & MDETR & 65.75 & 73.62 & 76.69 \\
& MDETR + Gold                  & 67.03 & 73.81 & 76.98 \\
\bottomrule[1.25pt]
\end{tabular}
}
\caption{
Results of different grounding objective, where for the pronoun (resp. noun phrase) task, the noun phrase (resp. pronoun) is not trained and evaluated.
}\label{table:objective}
\end{table}

\paragraph{Single Grounding Objective.}
Our extended task grounds the noun phrases and pronouns simultaneously, as illustrated in §\ref{sec:mainresult}.
We now evaluate the MDETR and \texttt{Gold} in single grounding objectives, i.e., ground only noun phrases or pronouns.
Table~\ref{table:objective} lists the test scores.
First, our suggested coreference graph encoding with \texttt{Gold} annotations could consistently boost performance in both sub-tasks.
Then, all methods in phrase grounding only task exists notable performance gap to our extended task results (Table~\ref{table:result}), e.g., in \textsc{Any-Box} recall@$1$, $39.94$ (extended) $\rightarrow 38.98$ (phrase only) of MDETR, $41.91 \rightarrow 41.43$ of \texttt{Gold}.
This shows that learning to ground both pronouns and noun phrases could promote the dialog and scenery understanding, thus improving the phrase grounding.

\begin{table}
\centering
\resizebox{0.44\textwidth}{!}{
\begin{tabular}{lccc}
\toprule[1.25pt]
& \multicolumn{3}{c}{Recall@1} \\ \cmidrule(){2-4}
Model & Overall & Pronoun & Phrase \\
\midrule
\multicolumn{4}{c}{\textsc{Any-Box-Protocol}} \\
\midrule
MDETR + Gold          & 47.54 & 58.79 & 41.91 \\
~~w/o \texttt{coref}  & 44.22 & 54.48 & 39.58 \\
~~w/o Virtual Span    & 46.21 & 55.61 & 41.52 \\
~~w/o \texttt{coref} \& Virtual Span  & 43.56 & 52.60 & 39.04 \\
\midrule[0.05pt]
MDETR               & 43.35 & 50.15 & 39.94 \\
\midrule
\multicolumn{4}{c}{\textsc{Merged-Box-Protocol}} \\
\midrule
MDETR + Gold          & 55.43 & 65.86 & 50.26 \\
~~w/o \texttt{coref}  & 52.71 & 63.74 & 47.19 \\
~~w/o Virtual Span    & 54.31 & 64.41 & 48.63 \\
~~w/o \texttt{coref} \& Virtual Span  & 51.96 & 62.46 & 46.71 \\
\midrule[0.05pt]
MDETR               & 51.98 & 60.86 & 47.59 \\
\bottomrule[1.25pt]
\end{tabular}
}
\caption{
Ablation study of graph encoding.
}\label{table:ablation}
\end{table}

\subsection{Ablation Study}\label{sec:ablation}
To verify the effectiveness of our designed coreference graph, we conduct ablation experiments in the \texttt{gold} coreference setting.

\paragraph{Coreference Edge.}
We first drop \texttt{coref} edges to show the importance of text coreference knowledge.
As presented in Table~\ref{table:ablation}, obviously, the model performance decrease dramatically in both protocols.
However, the graph with virtual spans provides mild improvements, we investigate this at the last.

\paragraph{Virtual Span Node.}
The virtual spans are used to represent the multi-word text mentions.
Here we remove them and the \texttt{span-word} \& \texttt{word-span} edges, then densely connect every words with each other in one coreference cluster by the \texttt{coref} edge.
As shown, the model performance is degraded to a certain extent.
Thus, the virtual span scheme is with not only conceptual advantages but also better performance.
In addition, we can directly use the span node features when applying to other span-based models, like \citet{liu-hockenmaier-2019-phrase}.

\paragraph{Only Word Node and Relation.}
In the end, we remove both \texttt{coref} and virtual span, keeping only \texttt{next-word}, \texttt{last-word}, and \texttt{self-loop} edges.
In Table~\ref{table:ablation}, we can see that this model is only comparable to the baseline.
First, this corroborates, without virtual spans, the coreference is still effective (the above paragraph).
Moreover, virtual span nodes alone act as the span indicator could improve the model as well (the first paragraph).

\section{Related work}

\paragraph{Visual Grounding.} General visual grounding, also known as referring expression comprehension \citep{transvg, RECsurvey}, is akin to phrase grounding to some extent, since they all aim to study the mapping from the expressions to the specific image regions. The main difference between them is that the visual grounding particularly focuses on one single expression, while the phrase grounding is more general and can be applied to multiple expressions.

\paragraph{Phrase Grounding.} A wealth of prior work \citep{DBLP:conf/mm/YuHYLYZL20,DBLP:conf/cvpr/DoganSG19,wang-etal-2020-maf,DBLP:conf/iccv/KamathSLSMC21} on phrase grounding has achieved promising results. Typically, \citet{G3raphGround} present an end-to-end framework with a separate graph neural network to explore phrase grounding, and \citet{graphforvisualgrounding} enhance this task by proposing a language-guided graph representation to capture the global context of grounding entities and their relations. In this work, we first propose that grounding pronouns is indispensable, then follow the foundation of using graph structures to our task, positing that the extra coreference knowledge in texts are positive and useful.

\paragraph{Visual Coreference Resolution.} It is true that our proposed task has some similarities with the visual coreference resolution task. \citet{yu-etal-2019-see} formalizes visual-aware pronoun coreference resolution (PCR), builds a dataset for PCR in visual-supported dialogues, and then presents a PCR model with image features. In other words, It solves the pronoun coreference at the text side with the help of visual information. In contrast, our task tackles coreference across the text and image, and in addition, we are also concerned about noun phrases, not only the pronouns. Additionally, \citet{DBLP:conf/eccv/KotturMPBR18} indeed presents visual coreference resolution (VCR) very similar to ours, with only a difference in the coreference direction (image-to-text v.s. ours text-to-image). As it targets visual question answering, the work does not build a dataset for VCR nor evaluate it. Moreover, it handles VCR at the sentence level for each question in the visual dialogue. In our work, we focus on VCR directly, with a released benchmark dataset, initial models as well as benchmark results.
\paragraph{Related Datasets.} The usual visual grounding datasets \citep{DBLP:conf/eccv/YuPYBB16}, RefCOCO, RefCOCO+ and RefCOCOg, only include one simple expression without pronouns.
There are several benchmark datasets \citep{DBLP:conf/eccv/LinMBHPRDZ14, DBLP:journals/ijcv/KrishnaZGJHKCKL17} for phrase grounding, and the most well-known is Flickr30k Entities dataset \citep{DBLP:conf/iccv/PlummerWCCHL15}. 
Nevertheless, since these datasets are among the first to build the relations between the noun phrases mentioned in a sentence and the specific localization of a corresponding image, 
they may ignore the pronouns, which can also be grounded and assistant to visual language understanding.

\section{Conclusion}

In this work, we proposed to extend phrase grounding task with pronouns, additionally, we established our dataset, \texttt{VD-Ref}, the first dataset which contains ground-truth mappings from both noun phrases and pronouns to image regions. Furthermore, we took the state-of-the-art model MDETR as our baseline and introduced extra coreference knowledge with graph neural networks. Experiments on our dataset showed the exciting phenomenon that pronouns are more accessible grounded than phrases and demonstrated the significance of coreference knowledge in visual language understanding. To this end, we conducted in-depth analyses of our results. In the future, we would expand more sophisticated dataset, and do more richer experiments on our dataset.

Our dataset and baseline code are avaliable at \href{https://github.com/izhx/Phrase-Grounding-with-Pronoun}{https://github.com/izhx/Phrase-Grounding-with-Pronoun}.

\section*{Limitations}
In this work, we collect our dataset and extend phrase grounding with pronouns by a series of explored experiments. Admittedly, due to the uneven distribution of raw data and complex annotation process, the main limitation is that our dataset only considers the visual phrases and pronouns, while lacking the annotations on non-visual textual expressions, and giving no insight into the scenery regions as well, which could restrict the research on more sophisticated conditions with varied coreference chains. Future work should be undertaken to expand a more complicated dataset and do more abundant experiments with coreference chains.

\section*{Ethical Statement}
We build the dataset \texttt{VD-Ref} to go on our researches, aiming to extend the phrase grounding task with pronouns, and study the performance where the coreference chains impact on.In the data annotation process, we adhere to a certain code of conduct on ethical consideration. When recruiting annotators for our task, we claim that all the potential annotators are free to choose whether they want to participate, and they can withdraw from the study anytime without any negative repercussions. Additionally, the whole annotation tasks are anonymized, totally agnostic to any private information of annotators.
Furthermore, the annotation results and dataset do not involve any sensitive information that may harm others. Overall, the establishment of our dataset is compliant with ethics.

\section*{Acknowledgments}
We thank all reviewers for their hard work.
This research is supported by grants from the Na-
tional Natural Science Foundation of China (No.
62176180).

\bibliography{custom}
\bibliographystyle{acl_natbib}

\appendix

\section{Annotation Interface}
\label{sec:appendix:interface}

\begin{figure}[H] 
  \centering
  \includegraphics[width=0.48\textwidth]{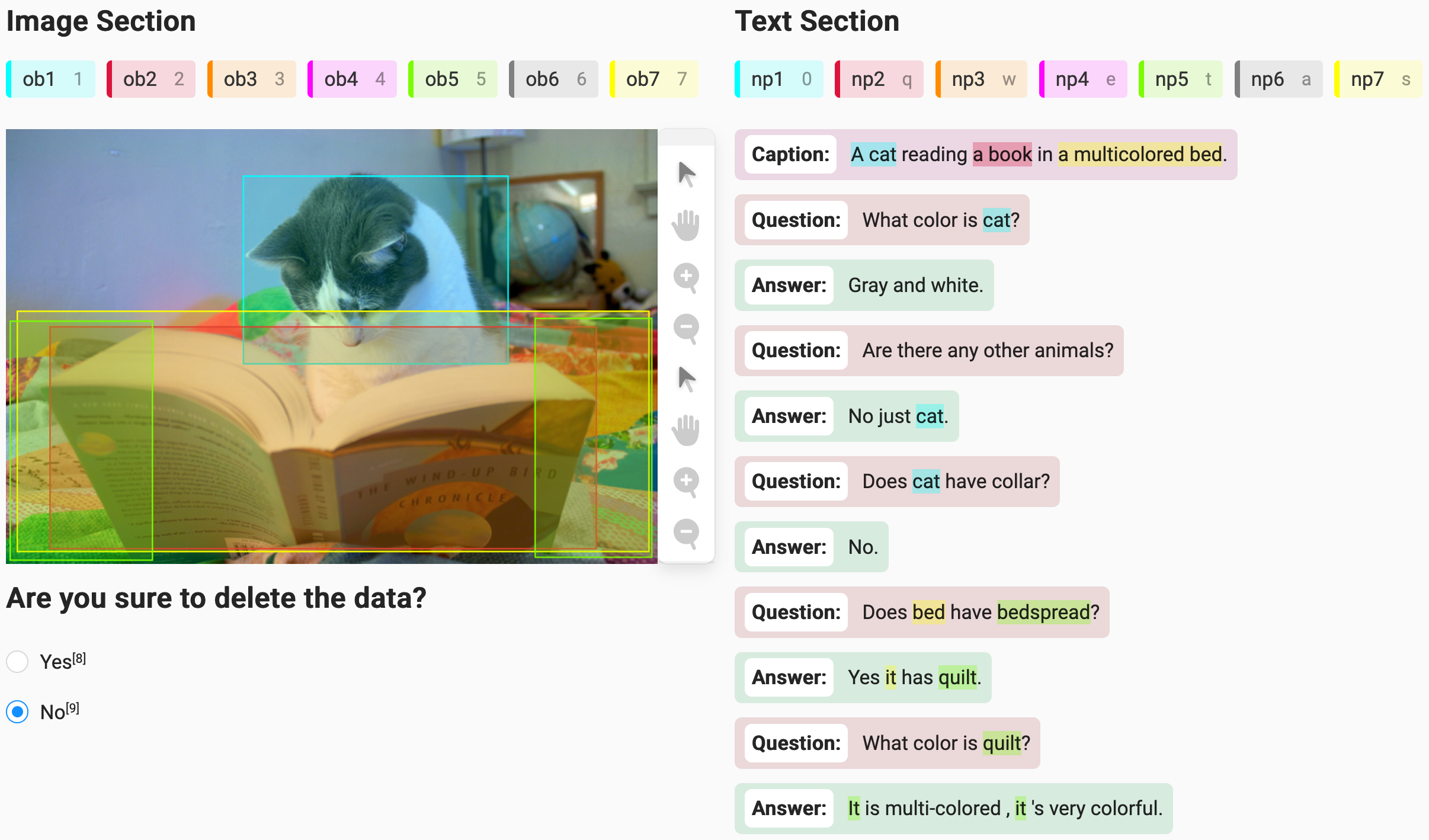}
  \caption{The designed interface of our annotation platform using label-studio tool.}
  \label{fig:interface}
\end{figure}

We designed the annotation interface. As illustrated in Figure \ref{fig:interface}, the left panel is the image section, while the right panel is the text section. They both have several boxes with distinct colors, which are used to annotate image regions and textual expressions. Moreover, the interface provides seven colors to choose from since the number of objects in the dialogue does not exceed 7 as a precondition. Notably, there is one option, ``Are you sure to delete the data?'', for the annotators and reviewers to remove vague and ambiguous datasets, where the dialogue contains too much irrelevant content or the image is incomplete, making it challenging to be recongnized.

\end{CJK}
\end{document}